\documentclass[10pt,twocolumn,letterpaper]{article}

\usepackage[pagenumbers]{cvpr} % To force page numbers, e.g. for an arXiv version
\usepackage{soul}
\usepackage{amsmath}
\usepackage{pifont}
\newcommand{\cmark}{\text{\ding{51}}}
\newcommand{\xmark}{\text{\ding{55}}}

\usepackage[dvipsnames]{xcolor}

\definecolor{cvprblue}{rgb}{0.21,0.49,0.74}
\usepackage[pagebackref,breaklinks,colorlinks,citecolor=cvprblue]{hyperref}

\def\paperID{284} % *** Enter the Paper ID here
\def\confName{3DV\xspace}
\def\confYear{2024\xspace}

\title{GAN-Avatar: Controllable Personalized GAN-based Human Head Avatar}

\author{
Berna Kabadayi\textsuperscript{1}
\and Wojciech Zielonka\textsuperscript{1}
\and Bharat Lal Bhatnagar\textsuperscript{2,3,5}
\and
\and Gerard Pons-Moll\textsuperscript{2,3}\qquad
Justus Thies\textsuperscript{1,4}
\and
\textsuperscript{1}Max Planck Institute for Intelligent Systems, T\"{u}bingen, Germany \\
\textsuperscript{2}Max Planck Institute for Informatics, Germany \\
\textsuperscript{3}University of T\"{u}bingen\qquad
\textsuperscript{4}Technical University of Darmstadt\qquad 
\textsuperscript{5}Meta Reality Labs\\
\rurl{ganavatar.github.io}
}

\begin{document}
\iftrue
    \newcommand\rurl[1]{\href{https://#1}{\nolinkurl{#1}}}
\else 

\fi

\crefname{figure}{Fig.}{Figs.}
\Crefname{figure}{Fig.}{Figs.}
\crefname{section}{Sec.}{Secs.}
\Crefname{section}{Sec.}{Secs.}
\Crefname{table}{Tab.}{Tabs.}
\crefname{table}{Tab.}{Tabs.}

\renewcommand{\paragraph}[1]{\smallskip\noindent\textbf{#1}}
\twocolumn[{%
\renewcommand\twocolumn[1][]{#1}%
\maketitle
\begin{center}
    \vspace*{-0.5cm}
    \centering
    \captionsetup{type=figure}
    \includegraphics[width=\linewidth, trim={0 8.1in 0 0},clip]
    {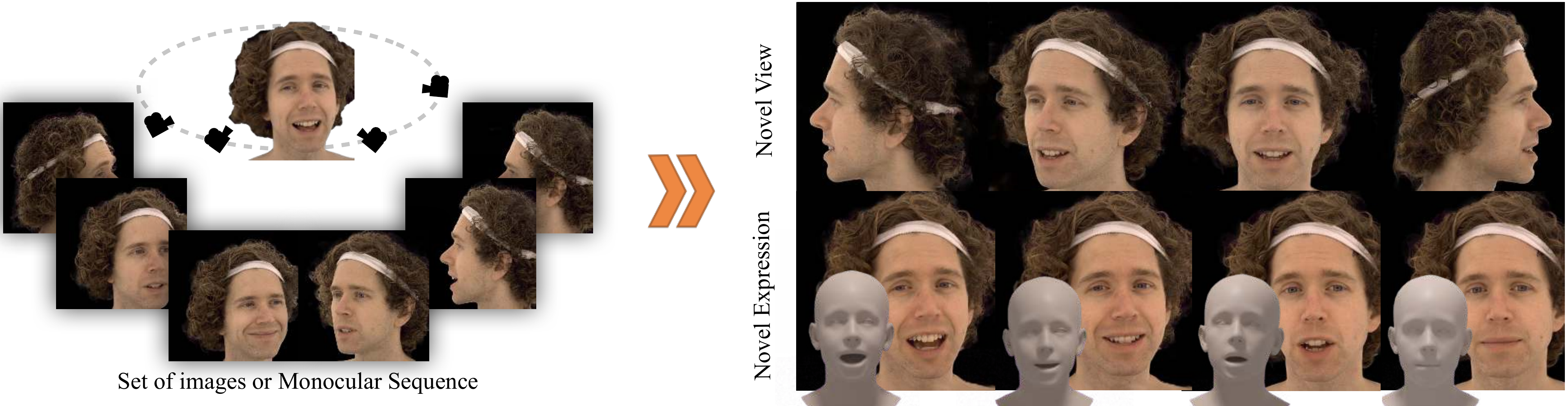} 
    \vspace{-0.4cm}
    \caption{
    Given a set of images of a person and the corresponding camera parameters, we construct an animatable 3D human head avatar.
    In contrast to previous work on personalized avatar reconstruction, we do not rely on precise tracking information of the facial expressions in the training data. 
    A generative adversarial network is trained to capture the appearance without facial expression supervision.
    To control the appearance model, we learn a mapping network that enables the traversal of the latent space by parametric face model parameters.
  }
    \label{fig:teaser}
\end{center}%
\vspace{0.2cm}
    }]

\begin{abstract}
Digital humans and, especially, 3D facial avatars have raised a lot of attention in the past years, as they are the backbone of several applications like immersive telepresence in AR or VR.
Despite the progress, facial avatars reconstructed from commodity hardware are incomplete and miss out on parts of the side and back of the head, severely limiting the usability of the avatar.
This limitation in prior work stems from their requirement of face tracking, which fails for profile and back views.
To address this issue, we propose to learn person-specific animatable avatars from images without assuming to have access to precise facial expression tracking.
At the core of our method, we leverage a 3D-aware generative model that is trained to reproduce the distribution of facial expressions from the training data.
To train this appearance model, we only assume to have a collection of 2D images with the corresponding camera parameters.
For controlling the model, we learn a mapping from 3DMM facial expression parameters to the latent space of the generative model.
This mapping can be learned by sampling the latent space of the appearance model and reconstructing the facial parameters from a normalized frontal view, where facial expression estimation performs well.
With this scheme, we decouple 3D appearance reconstruction and animation control to achieve high fidelity in image synthesis.
In a series of experiments, we compare our proposed technique to state-of-the-art monocular methods and show superior quality while not requiring expression tracking of the training data.

\end{abstract}    
\section{Introduction}
\begin{figure}[t]
    \centering
    \includegraphics[width=0.7\linewidth]{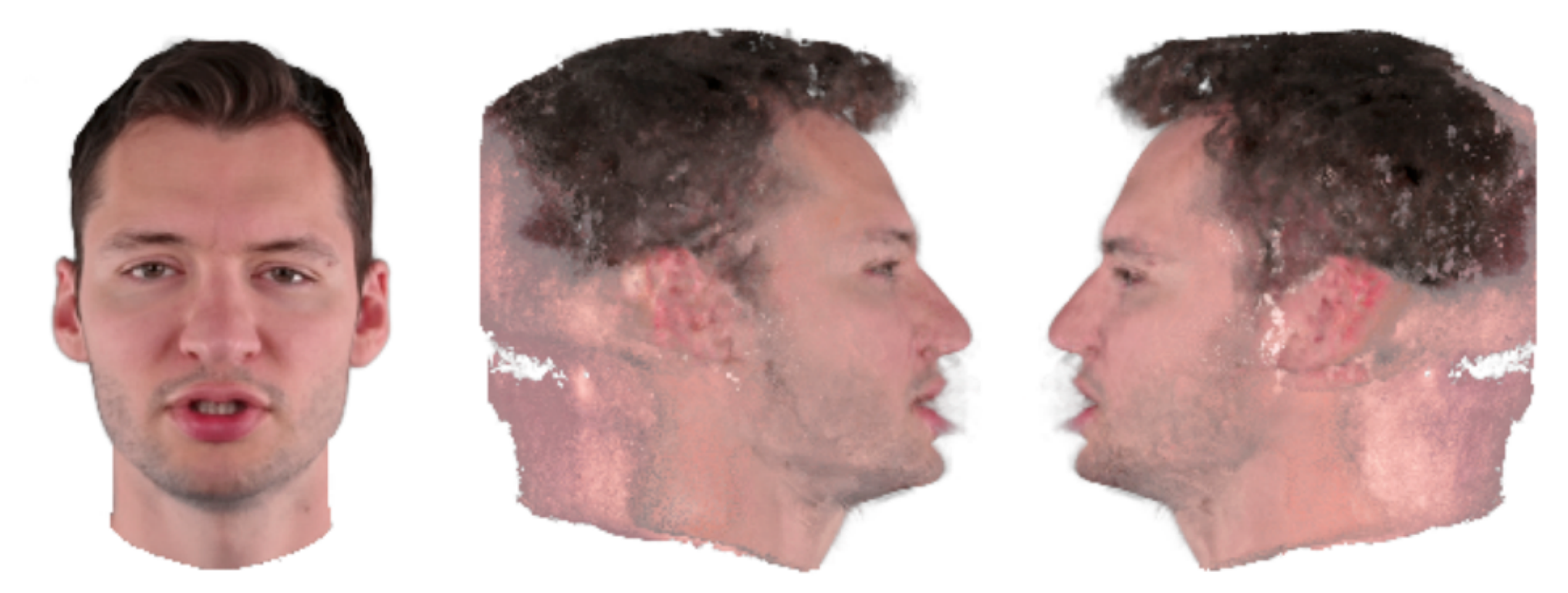}
    \caption{
    Since monocular 3D avatar methods like INSTA~\cite{zielonka2023instant} rely on facial expression tracking for the employed reconstruction losses, they cannot reconstruct a complete head avatar, including the back or sides of the head, as face tracking fails on those views.
    %s
    }
    \label{fig:tracking_issues}
\end{figure}

In recent years, we have seen immense progress in digitizing humans for applications in augmented or virtual reality.
Digital humans are the backbone of immersive telepresence (e.g., metaverse), as well as for many entertainment applications (e.g., video game characters), movie editing (i.e., special effects, virtual dubbing), and e-commerce (e.g., virtual mirrors, person-specific clothing).
For these use cases, we require complete reconstructions of the human head to allow for novel viewpoint synthesis.
Recent methods to recover an animatable digital double of a person either use monocular~\cite{zielonka2023instant, Zheng:CVPR:2022, grassal2022neural,Gafni_2021_CVPR, chen2023implicithead, bai2023learning, feng2023learning, bhatnagar2019mgn, alldieck2019learning, xue2023nsf, xie22chore} or multi-view inputs~\cite{wuu2022multiface, Lombardi21, Cao2022AuthenticVA, hq3davatar2023, Li2023MEGANEME, Wang2022MoRFMR, Kirschstein2023NeRSembleMR, bhatnagar2020ipnet, bhatnagar2020loopreg, bhatnagar22behave}.
The appeal of monocular approaches is the wide applicability, as anyone can record the input data using a webcam or smartphone.
As a prior, those methods rely on parametric face models like FLAME~\cite{flame} or BFM~\cite{bfm} to control the 3D avatar.
Recent learning-based monocular approaches are IMavatar~\cite{Zheng:CVPR:2022}, INSTA~\cite{zielonka2023instant}, NerFace~\cite{Gafni_2021_CVPR}, NHA~\cite{grassal2022neural}.
Although monocular approaches are handy to reconstruct, they heavily rely on precise face tracking during training.
Oftentimes, their accuracy is limited by the 3D facial expression tracker and the underlying detection of facial landmarks used to train face regressors ~\cite{deca, Deng2019Accurate3F} or during optimization~\cite{face2face, Garrido2013ReconstructingDD, Valgaerts2012LightweightBF}.
3D tracking is hard~\cite{deca, Deng2019Accurate3F, xie2023vistracker, zhou2022toch, face2face}, and when landmark detection fails, these methods will likely also fail.
This happens for profile views or when the person looks away from the camera.
Thus, recent monocular methods are limited to the frontal appearance and do not include the back of the head; see \Cref{fig:tracking_issues}.
Reconstructing personalized head avatars through the use of a multi-view setup can be used instead.
The complexity of such setups can vary widely, from using just a couple of DSLR cameras~\cite{Beeler2010HighqualitySC} to setting up an expensive camera dome \cite{wuu2022multiface} with dozens of cameras and controllable light~\cite{Wenger2005PerformanceRA, lightstage, relightables}.
Highly detailed faces captured in such studios serve many purposes in various areas, from the gaming industry to visual effects in movies and games, or for collecting training data.
However, they are expensive and not accessible to everyone.
Similar to recent monocular methods, multi-view methods~\cite {Lombardi21, Li2023MEGANEME, Cao2022AuthenticVA} also rely on precise tracking of the face (e.g., based on template tracking). 
Thus, both monocular and multi-view approaches, are bound by the quality of the facial expression tracking.
In contrast, the goal of this work is to reconstruct a complete head avatar without relying on precise facial expression tracking information.
Specifically, we construct an appearance model using image data, where only the corresponding camera parameters are available, and per-frame geometry is \textit{not} needed.
We do not rely on any predictions like semantic face parsing~\cite{zielonka2023instant,grassal2022neural} or predicted normal maps~\cite{grassal2022neural} as done in state-of-the-art monocular avatar reconstruction methods (see \Cref{tab:inputcorpus}).
At the core of our method is a 3D-aware generative appearance model, which leverages a pre-trained EG3D~\cite{chan2022efficient} model.
Using the known camera parameters of the input dataset of a person, we fine-tune the appearance model to match the distribution of the observations.
This yields us a personalized 3D appearance model.
To control this appearance model with standard expression parameters of the BFM model~\cite{bfm}, we devise a mapping network that maps expression codes to latent codes of the generative model.
To this end, we sample the generator and render the facial appearance in a normalized, frontal view where facial expression estimation works reliably and train the mapping network in a supervised fashion.
In our experiments, we show that our idea of decoupling appearance reconstruction and controllability leads to high-quality head avatars without the requirement of precise facial expression tracking of the input training data.
As a result, we achieve sharper appearances compared to state-of-the-art methods, particularly in teeth and hair regions.

\medskip
\noindent
In summary, we propose the following contributions:
% itemize
\begin{itemize}
    \item a generative 3D-aware person-specific head avatar appearance model that can be trained without the need for precise facial expression tracking,
    \item and an expression mapping network that gives control over the model, allowing us to generate novel animations under novel views.
\end{itemize}
\section{Related Work}
Our method learns a personalized facial avatar of a subject by combining a generative 3D-aware model with a facial expression mapping network.
In the following, we review the state-of-the-art for 3D head avatar reconstruction methods and generative face models.

% Table
\begin{table}[t]%
    \centering    
    \resizebox{\linewidth}{!}{
    \begin{tabular}{ccccc}
      \toprule
      Method   & Cam. & Facial Expr. & Mesh & Seg. mask \\ 
      \midrule
      IMavatar \cite{Zheng:CVPR:2022}     & \cmark & \cmark & \cmark & \xmark \\
      NeRFace \cite{Gafni_2021_CVPR}      & \cmark & \cmark & \xmark & \xmark \\
      INSTA \cite{zielonka2023instant}     & \cmark & \cmark & \cmark & \cmark \\
      \midrule
      Our Method    &    \cmark &  \xmark &  \xmark &  \xmark \\
      \bottomrule
    \end{tabular}
    }
    \caption{
    Training corpus requirements of state-of-the-art monocular avatar reconstruction methods. 
    In contrast to methods that require inputs like per-frame facial expressions, guiding mesh reconstructions, or semantic facial parsing masks, our proposed method only requires the camera parameters to learn a personalized avatar.%appearance model.
    }
    \label{tab:inputcorpus}
\end{table}%

\paragraph{Monocular Head Avatar Reconstruction}
Since estimating 3D face geometry from 2D images has many none-face-like solutions, a strong geometric prior is needed.
Therefore, most state-of-the-art methods use parametric face models~\cite{morphable} like FLAME~\cite{flame} to stay in a plausible solution space.
INSTA~\cite{zielonka2023instant} uses the metrical face tracker from MICA~\cite{Zielonka2022TowardsMR} to estimate per-frame FLAME~\cite{flame} parameters and embeds a neural dynamic radiance field (NeRF)~\cite{Mildenhall2020NeRFRS} around the 3D mesh.
The triangles of the mesh create local transformation gradients used for the projection of points sampled on the ray between canonical and deformed spaces~\cite{Pumarola2020DNeRFNR}.
Thus, INSTA~\cite{zielonka2023instant} relies heavily on precise tracking without a mechanism to compensate for tracking failures.
IMavatar~\cite{Zheng:CVPR:2022} uses face tracking from DECA~\cite{deca} as initialization and refines poses and expression parameters during appearance learning.
It uses coordinate neural networks to span 3D skinning weights, which are used to deform the volume~\cite{Chen2021SNARFDF, Saito2021SCANimateWS}.
Similar to INSTA~\cite{zielonka2023instant}, it requires a good tracking initialization and needs to be trained for several days for a single subject.
PointAvatar \cite{Zheng2022PointAvatarDP} is a deformable point-based method that tackles the problem of efficient rendering and reconstruction of head avatars with a focus on thin structures like hair strands.
Except for using point cloud representations, the other main difference to IMavatar \cite{Zheng:CVPR:2022} is a single forward pass for the optimization and rendering, eliminating the heavy root-finding procedure for correspondence search between points in the canonical and deformed spaces.
Unfortunately, the point-based formulation exhibits holes in the avatars, thus, lowering the visual quality.
Moreover, all the above methods rely on tracked meshes for additional geometry regularization.
In contrast to INSTA~\cite{zielonka2023instant}, IMavatar~\cite{Zheng:CVPR:2022}, or PointAvatar~\cite{Zheng2022PointAvatarDP}, NeRFace~\cite{Gafni_2021_CVPR} does not use a canonical space to model the appearance of a subject, but directly operates in the posed space using an MLP which is conditioned on facial expression parameters~\cite{bfm, face2face}. % and outputs the radiance and densities of the posed face.
NeRFace~\cite{Gafni_2021_CVPR} tends to overfit the training data and fails to render novel views.

NHA \cite{grassal2022neural} is an avatar method that uses an explicit representation for the geometry, i.e., a mesh based on FLAME.
It uses a face tracking scheme following Face2Face~\cite{face2face} and optimizes for expression-dependent displacements and a neural texture~\cite{thies2019neural} to reproduce the appearance.
Similar to NeRFace, it fails to render novel views correctly and often exhibits geometrical artifacts for ears~\cite{zielonka2023instant}.

\paragraph{Multi-view Head Avatar Reconstruction}
For high-quality head avatar reconstruction, calibrated multi-view setups are used.
They enable precise face tracking using optimization-based reconstruction~\cite{alexander2009,beeler2011} or learned tracking~\cite{laine2017tracking} which can be used to guide learned appearance representations.
MVP~\cite{Lombardi21} allocates voxels called volumetric primitives on the vertices of the meshes captured in a high-end multi-view camera dome.
Each of the primitives is allowed to deviate from the initial position. Additionally, the voxels store payloads of alpha and RGB values which are optimized using volumetric rendering~\cite{Lombardi:2019}.
Despite the excellent quality and the ability to capture a vast amount of materials, the method requires personalized face tracking~\cite{laine2017tracking}.
Pixel Codec Avatars (PiCa)~\cite{Ma2021PixelCA} is another approach heavily relying on preprocessed geometry.
Similarly to MVP~\cite{Lombardi21} the method is based on an encoder-decoder architecture.
An avatar codec is computed using a convolutional neural network which takes the per-frame mesh (unwrapped into a position map using a UV parametrization) and the average texture as input.
From this codec, the position map and local appearance codes can be decoded, which are used for a per-pixel decoding to compute the final image.
The whole process is supervised by tracked meshes and depth maps.
In order to generalize MVP to multiple subjects, Cao et al.~\cite{Cao2022AuthenticVA} introduced a cross-identity hyper network (identity encoder) that requires a few phone scans as input in order for the method to produce high-quality avatars.
Given a user's average texture and geometry, the hypernetwork predicts a set of multiscale bias maps per subject.
Those maps are later used to condition the MVP's decoder to render an image.
In contrast to those multi-view-based avatar reconstruction methods, our proposed method can be applied to monocular data and more importantly, does not require precise geometry tracking for training the appearance model.

\begin{figure*}[t]
    \centering
    \includegraphics[width=\linewidth, trim={1.3in 7.2in 1.3in 0},clip]
    {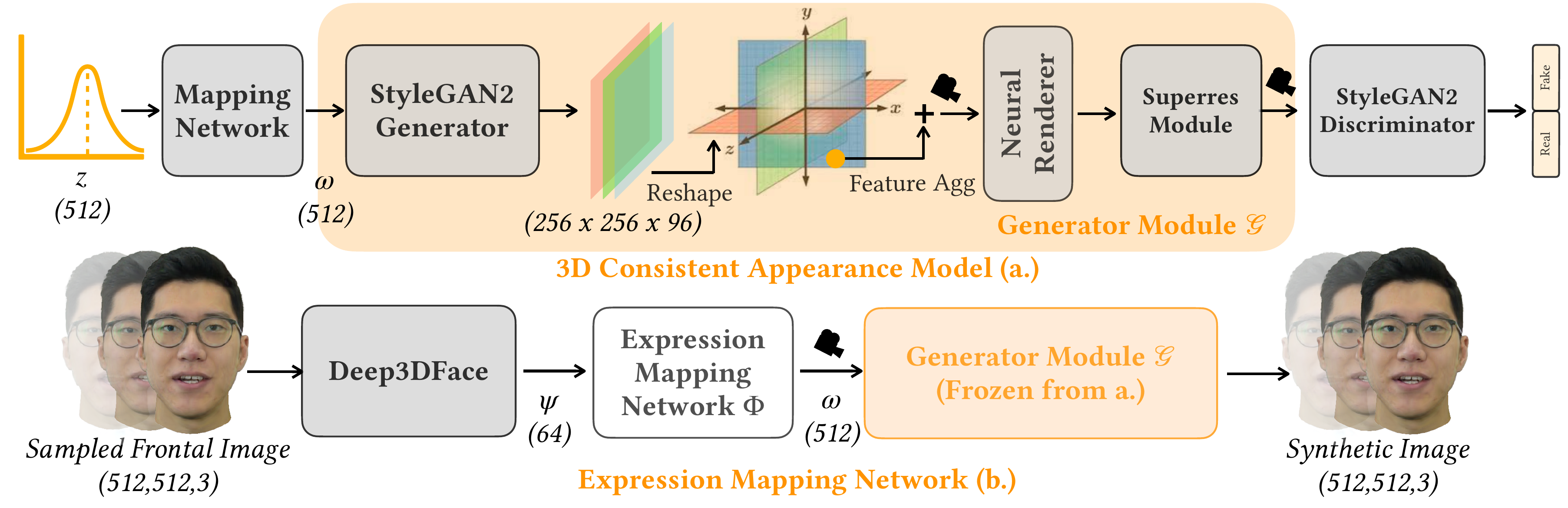}
    \caption{
    Method overview. For our actors, we fine-tune EG3D \cite{chan2022efficient} trained on FFHQ. Compared to the original EG3D, only our discriminator knows the camera pose $\mathbf{p}$. (b) From frontal-looking images (easy to reconstruct) generated from the model, we regress facial expression parameters $\psi$ using Deep3DFace \cite{Deng2019Accurate3F}. Our expression mapping network $\Phi_\Theta(\psi)$ predicts the learned latent code $\omega$ from an input expression code $\psi$. For an expected $\omega$ code, using the generator module $\mathcal{G}$, we render the image and minimize the photometric loss between the rendered image and the fake input image. The generator module $\mathcal{G}$ is frozen while training the mapping network.
    }
    \label{fig:architecture}
\end{figure*}

\paragraph{3D-Aware Generative Models for Faces}
StyleGAN~\cite{Karras2018ASG} and its numerous follow-up works~\cite{Karras2019AnalyzingAI, Karras2021AliasFreeGA} are able to generate high-quality 2D images of human faces using a progressive GAN training scheme.
It has been extended to 3D-aware generative models.
Pi-Gan~\cite{Chan2021pigan} was one of the first methods which combined generative color and geometry.
Based on a NeRF-based volumetric rendering and a StyleGAN mapping network with FiLM conditioning~\cite{Perez2017FiLMVR} that is adapted to utilize sinusoidal activation functions \cite{Sitzmann2020ImplicitNR}, pi-GAN can sample high-quality images.
However, the generated proxy geometry is low quality, and the generated images are not multi-view consistent.
EG3D~\cite{chan2022efficient} explicitly targets those shortcomings.
It uses the StyleGAN generator to predict three feature maps, interpreted as a low-dimensional approximation of a 3D volume (tri-plane representation).
For each 3D point, a feature vector is calculated by projecting it onto each of the three feature planes to be later decoded by the downstream NeRF renderer.
Finally, the StyleGAN discriminator is used as a loss function.
LatentAvatar \cite{xu2023latentavatar} uses an image as conditioning to generate the triplane feature maps.
Despite high-quality rendering of frontal images,  EG3D struggles to produce $360^{\circ}$ views because it is trained on mostly frontal images where face detection and landmark predictors work, which are needed to normalize the data.
To address this problem, PanoHead~\cite{An2023PanoHeadG3} extends the training corpus of EG3D by carefully capturing data from the sides and the back of the head and replaces the tri-planes with grids.
%
%In addition, PanoHead replaces the tri-planes with grids, introducing a tri-grid that naturally allows the inclusion of depth in the encoding.
%
The 3D-aware GANs listed above can be used to generate novel people or to reconstruct a 3D model from an image using GAN inversion~\cite{lin20223d, ganinversion}. 
Recent methods extend EG3D to also incorporate expression control~\cite{yue2022anifacegan, 3dfaceshop}. However, the animation of such an avatar is uncanny as facial details like teeth change from frame to frame. 

\section{Method}
Given a set of images of a speaking person with the corresponding camera parameters, we aim to reconstruct an animatable, 3D-consistent human head avatar.
In contrast to previous work, we propose a method that does not require facial expression tracking of the training data to construct an appearance model.
Specifically, we devise a generative model based on EG3D \cite{chan2022efficient} to learn a person-specific appearance and geometry.
By leveraging a pre-trained model based on the FFHQ~\cite{Karras2018ASG}, we bootstrap our model to have fast convergence and diverse facial expressions.
Once the appearance model is trained on the input data, we generate training data for a mapping network that enables animation by mapping BFM expression parameters to the latent space ($\boldsymbol{\mathcal{W}}$ space) of the GAN model \cite{tewari2020stylerig, bermano2022stateoftheart}.
We render normalized images of the subject by sampling the generative appearance model and reconstruct the facial expression parameters for the individual images using \cite{Deng2019Accurate3F}.
Note that in contrast to the input images, the facial expressions in the sampled images are more straightforward to reconstruct as they are rendered in a frontal orientation, without side views, where face reconstruction methods struggle.
Using these samples with latent code and expression pairs, expression mapping is learned.
In the following, we will detail our proposed method, which is also depicted in \Cref{fig:architecture}.

\begin{figure*}[t]
    \centering
    \includegraphics[keepaspectratio=true, width=\linewidth, height=0.20\linewidth]{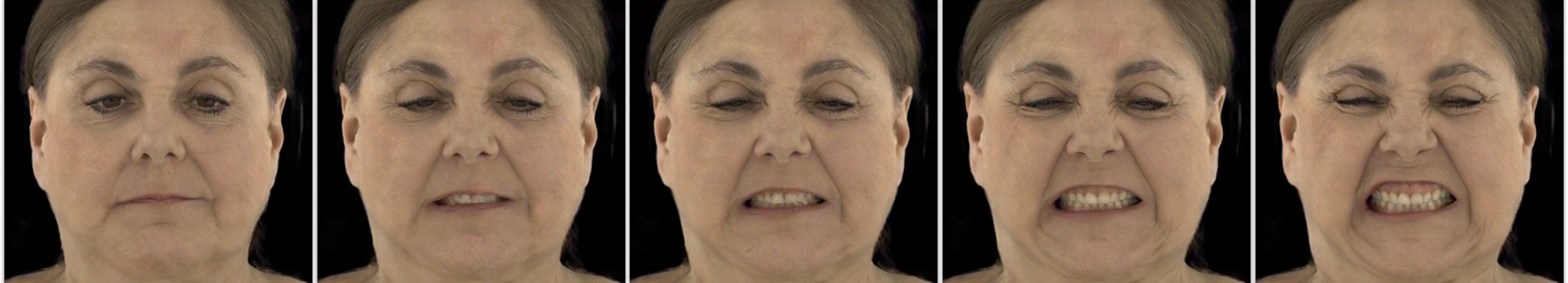}
    \caption{Linear latent space interpolation between two keyframes (left and rightmost). Our person-specific generative model has a well-shaped latent space which allows for a smooth interpolation between expressions. Actor from the Multiface dataset~\cite{wuu2022multiface}.
    }
    \label{fig:interp}
\end{figure*}

\subsection{3D-Consistent Appearance Model}
\label{sec:appearance_model}
As a backbone for the 3D-consistent appearance model, we use the efficient EG3D~\cite{chan2022efficient} tri-plane representation.
It leverages a StyleGAN2~\cite{Karras2019AnalyzingAI} architecture to generate the three feature planes from a random latent code.
StyleGAN2 architecture includes mapping and synthesis networks. 
First, the StyleGAN2 mapping network learns a latent code $\omega \in \mathbb{R}^{512}$ from a given random latent code $z \in \mathbb{R}^{512}$. 
Second, the synthesis network generates a photorealistic image from learned $\omega$. 
In our case, instead of generating an image, following EG3D~\cite{chan2022efficient} architecture, we generate three triplanes from a learned $\omega$. 
These triplane features are then rendered using volumetric rendering. 
Within the 3D-consistent appearance model, we define a Generator Module $\mathcal{G}$, which generates an image ${I}_{gen}$:
\begin{equation}
\mathcal{I}_{gen} = {\mathcal{G}(\omega, \mathbf{p})},
\enspace
\end{equation}
where $\omega$ and $\mathbf{p}$ are learned latent code and camera parameters, respectively. 
The camera parameter $\mathbf{p}=(R,t,\mathcal{K})$ describes rotation $R\in SO(3)$, translation $t\in\mathbb{R}^{3}$ and intrinsics $\mathcal{K} \in \mathbb{R}^{3x3}$; see \Cref{fig:architecture}.

While the original EG3D \cite{chan2022efficient} is trained to generate different identities with different expressions and poses, we aim at a personalized model that captures all idiosyncrasies of the subject's head, including teeth and hair.
To this end, we train our method assuming a collection of 2D images of a single subject and the corresponding camera parameters.
%.

%
Instead of training the model from scratch, we initialize the network with weights from a general EG3D \cite{chan2022efficient} model trained on the FFHQ dataset~\cite{Karras2018ASG}.
To reuse these weights, we align the pre-trained EG3D \cite{chan2022efficient} model with our person-specific input images.
Specifically, we extract the geometry of a sampled face of the pre-trained model and apply (non-rigid) Procrustes to align the mesh with a reconstructed face from \textit{one} of the input images.
The resulting rotation, translation, and scale are applied globally to all camera parameters of the input.
In contrast to EG3D, we do not assume normalized camera parameters and images.
Instead, we adapt the rendering formulation using a ray-bounding box intersection test to place samples along the viewing rays around the head center.% during training and inference.
Using the pre-trained EG3D \cite{chan2022efficient} model allows us to leverage the large FFHQ dataset (70k images) statistics which include different expressions.
Specifically, we avoid GAN training issues when training the personalized model, such as mode collapse, leading to a less expressive appearance model.
Starting from the pre-trained model, we fine-tune the personalized, unconditional generative model for $300k$ steps for monocular sequences and $\sim{1M}$ steps for $360^{\circ}$ head experiments using the original StyleGAN2/EG3D loss formulations. 
We refer to~\Cref{sec:mvcomparisonmultiface} for hyperparameters used in the appearance model training.
In contrast to EG3D \cite{chan2022efficient}, we do not provide the camera parameters to the StyleGAN2 \cite{Karras2019AnalyzingAI} mapping network to avoid 3D inconsistencies.
We perform volume rendering at a resolution of $128^2$, and increase the number of samples for both coarse and fine sampling from $48$ to $120$.
Note that by fine-tuning the model to our input data, we force the GAN to learn the distribution of different facial expressions for a specific subject — it is not generating different people anymore.
In \Cref{fig:interp}, we show an interpolation in the latent space of such an appearance model.
As we can see, the model's latent space is well-behaved and results in smooth transitions between sampled expressions.

\subsection{Expression Mapping Network}

The 3D-consistent appearance model allows us to generate images of the subject from a predefined camera view. However, the controllability is missing.
To learn a mapping from classical facial expression codes (e.g., blend shape coefficients) to the latent codes, we generate paired data by sampling the GAN space similar to \cite{tewari2020stylerig}.
Given random latent codes $\omega$, we render $1000$ frontal-looking face images $\mathcal{I}_{gen}$ using our appearance model. %$W+$
%
% $$ \mathbb{R}\in \left[a_1,a_n\right] $$
We extract the expression parameters $\psi \in \mathbb{R}^{64}$ from these generated images by reconstructing a 3D face model using Deep3DFace~\cite{Deng2019Accurate3F}.
Note that the face reconstruction works reliably in these frontal views, in contrast to side and back views in the training data.
Potentially, a multi-view reconstruction method can be applied in future work, as the appearance model can be used to render many arbitrary views for a specific latent code.
The mapping network $\Phi_\Theta(\psi)$ is constructed to map the expression codes to the $\boldsymbol{\mathcal{W}}$ space of the StyleGAN2 network. 
Specifically, our expression mapping network data $\mathcal{D}$ consists of expression-latent pairs $(\psi,\omega)\in\mathcal{D}$. 
The network is trained to generate $\omega'=\Phi_\Theta(\psi)$, reproducing the image using a frozen Generator Module $\mathcal{G}(\omega', \mathbf{p})$ from a frontal camera $\mathbf{p}$ based on a photometric distance loss:
\begin{equation}
\mathcal{L}_{\text{pho}}(\Theta) = 
%\frac{1}{N_\mathcal{P}}
\sum_{(\psi,\omega )\in\mathcal{D}}
{
\big|\big| \mathcal{G}(\omega, \mathbf{p}) - \mathcal{G}(\Phi_\Theta(\psi), \mathbf{p}) \big|\big|_2^2}
\enspace . % \\[20pt]
\end{equation}

Our shallow expression mapping network is a multi-layer perceptron (MLP) which consists of $2$ hidden layers with ReLU activation, and a final linear output layer. 
The input and the hidden layer size is $64$, and the output size is $512$, which is the dimension of the learned latent vector of the generative appearance model.
We train our model $\sim 1k$ steps with AdamW \cite{adamW} using a learning rate of $0.0005$.

\begin{figure*}[t]
    \centering
    \includegraphics[keepaspectratio=true, width=\linewidth, height=\linewidth]{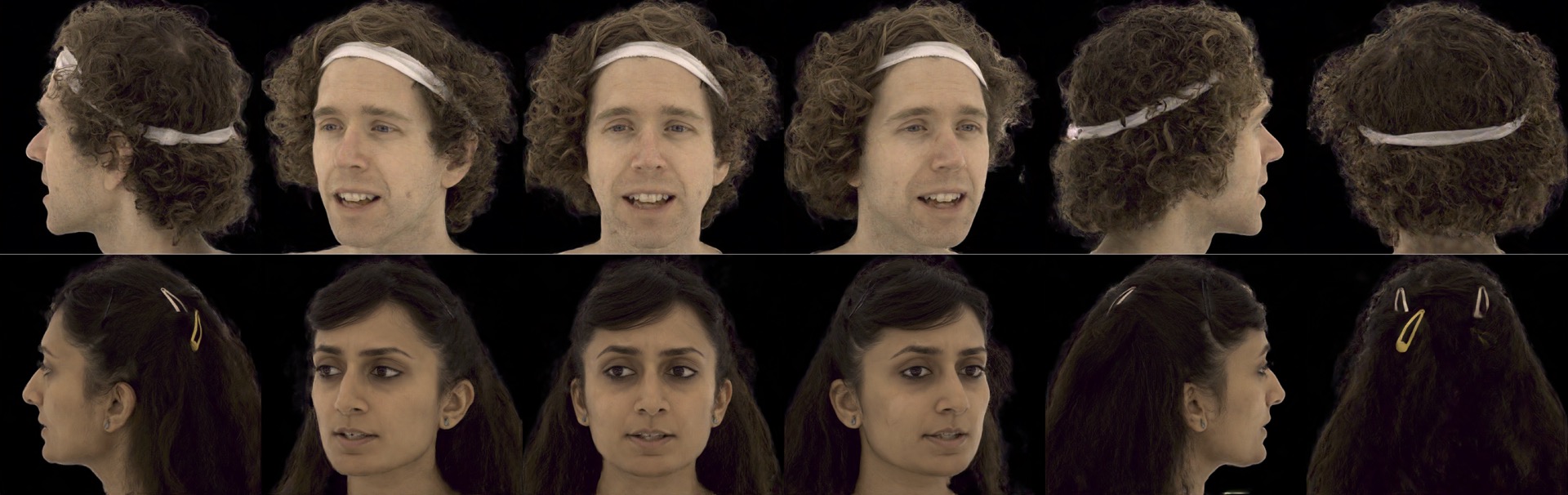}
    \vspace{-0.5cm}
    \caption{
        Our method synthesizes 3D-consistent novel views for full $360^\circ$ human head avatars which are animatable by facial expression parameters.
        To learn this avatar, we do not require facial expression tracking of the training sequence of the subject, thus resulting in a high-quality $360^\circ$  appearance.
        Actors are from the Multiface dataset~\cite{wuu2022multiface}.
    }
    \label{fig:360}
\end{figure*}

\section{Results}
\label{sec:results}
Unlike prior work, we build a 3D avatar of a person without relying on detailed 3D facial template tracking.
In the following, we analyze our method both qualitatively and quantitatively on monocular 
and multi-view data (see \Cref{sec:dataset}).
Specifically, we compare our approach with the state-of-the-art monocular avatar generation methods IMavatar~\cite{Zheng:CVPR:2022}, NerFace~\cite{Gafni_2021_CVPR} and INSTA~\cite{zielonka2023instant} in \Cref{sec:comparisonsota}, and provide ablation studies in \Cref{sec:ablation}.
\begin{figure*}[t]
\centering
    \includegraphics[width=\linewidth]{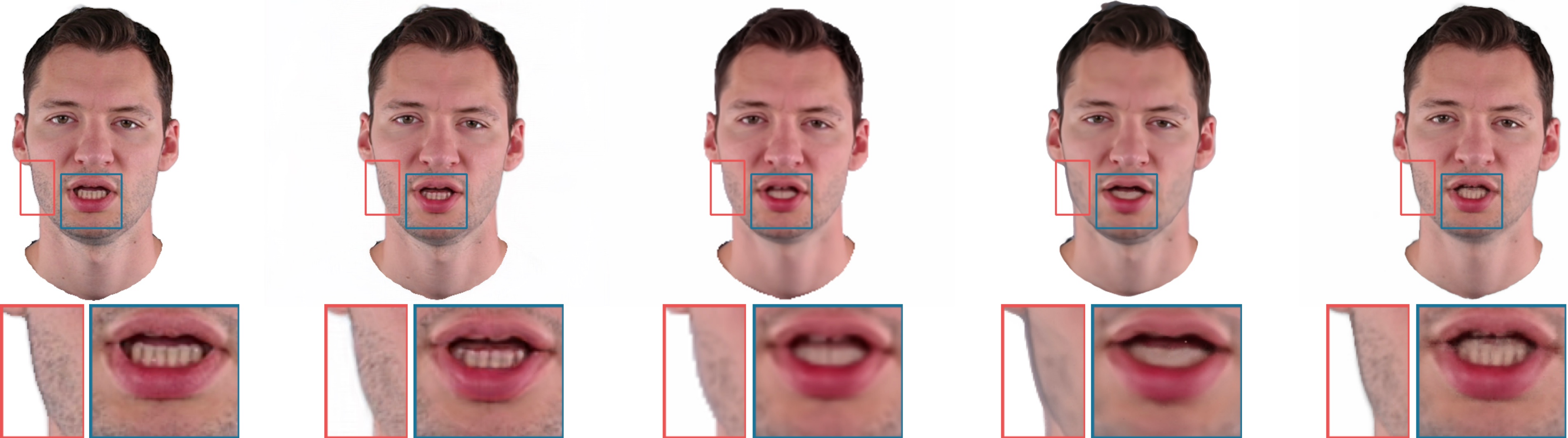} 
    \vspace{0.4em}
    \includegraphics[width=\linewidth]{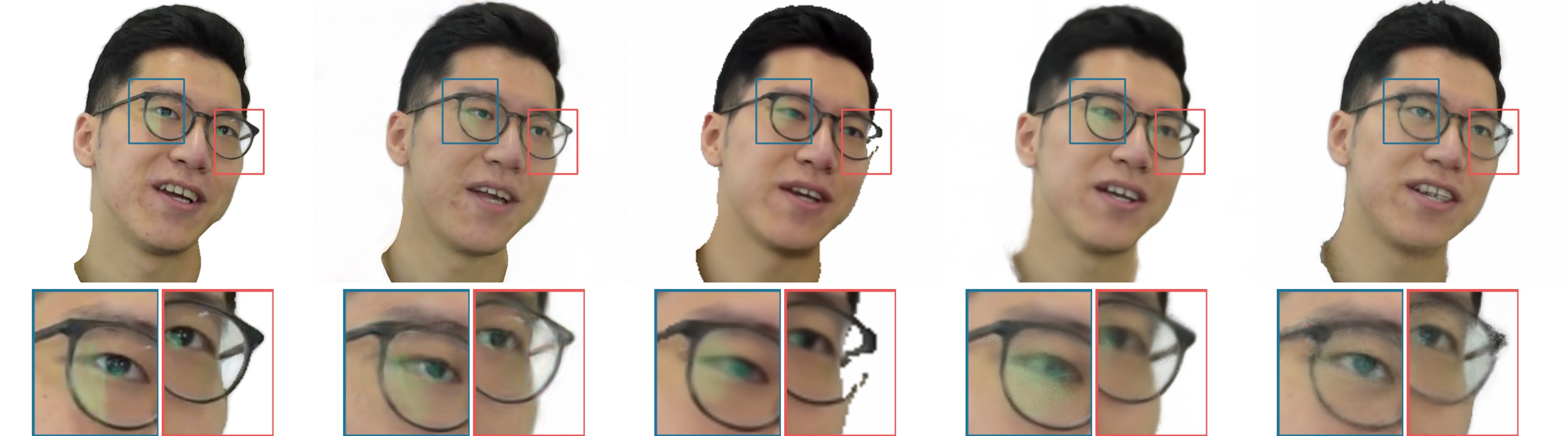} 
    \vspace{0.4em}
    \includegraphics[width=\linewidth]{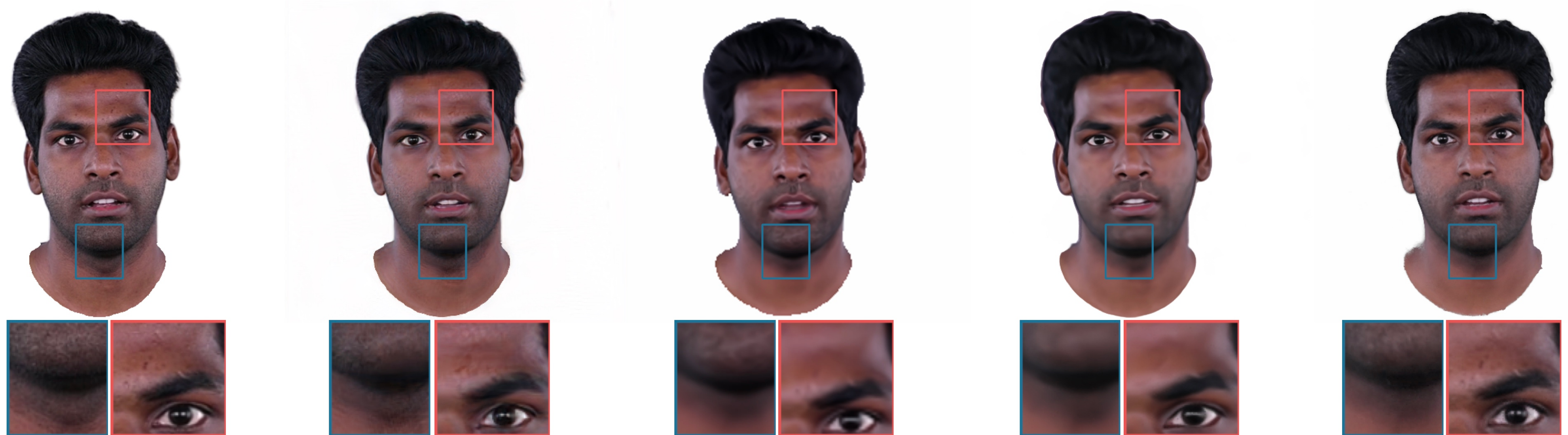}
    \vspace{0.4em}
    \includegraphics[width=\linewidth]{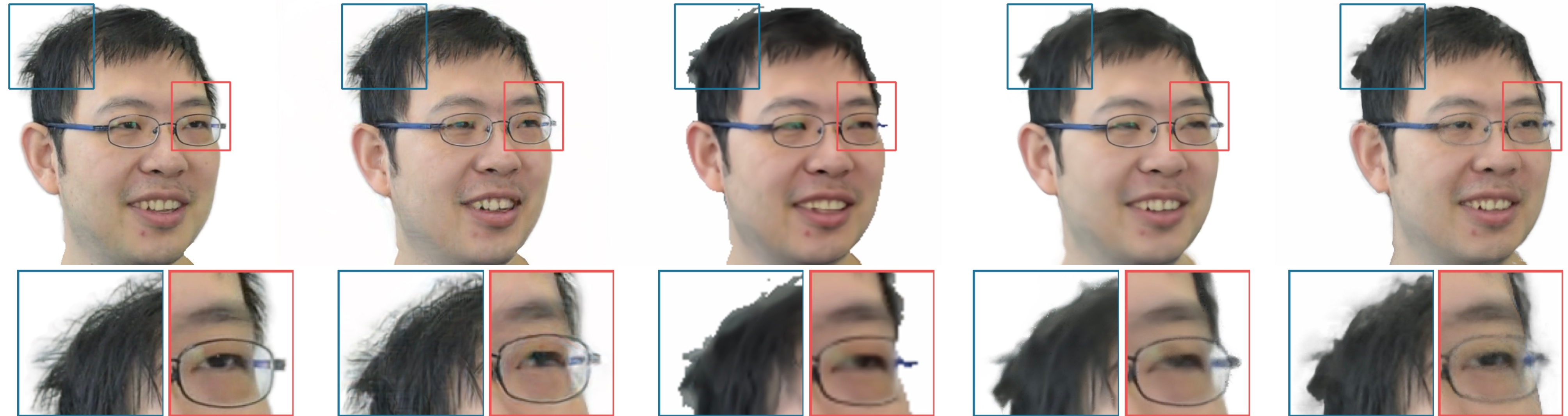} 
    \vspace{-0.75em}
    \newline
    \text{Ground Truth \hspace{1.7cm} Our Method  \hspace{1.7cm} IMavatar \cite{Zheng:CVPR:2022} \hspace{1.7cm} NeRFace \cite{Gafni_2021_CVPR}  \hspace{1.7cm} INSTA \cite{zielonka2023instant}}
     \caption{
     Our method can synthesize thin structures (e.g., hair strands) and a sharper texture, including teeth and skin compared to state-of-the-art monocular avatar methods.
     Actors are from the NeRFace~\cite{Gafni_2021_CVPR} and INSTA datasets~\cite{zielonka2023instant}.
     } 
    
    \label{fig:novelexpressionmonocular}
\end{figure*}

\subsection{Dataset and Evaluation Metrics}
\label{sec:dataset}

Our method takes images and the corresponding camera parameters as input to generate a full-head volumetric avatar.
We evaluate our method on two sets of data sources: monocular and multi-view data.

\paragraph{Monocular data} is taken from the publicly available datasets of NeRFace~\cite{Gafni_2021_CVPR} and INSTA~\cite{zielonka2023instant}, which are 2–3 mins long, recorded at a resolution of 25fps.
Following the evaluations in the baseline publications, the last 350 frames of the monocular videos are used for testing.

\paragraph{Multi-view experiments} are conducted on the publicly available actors from the Multiface v2 dataset covering the $360^{\circ}$ head ~\cite{wuu2022multiface} to evaluate the novel viewpoint synthesis and animation generation.
We pick 4-5 expressions from every actor, which we later crop and adjust to a $512\times512$ resolution.
We remove the background of the images using the image matting method of Lin et al.~\cite{videomatting} and apply gamma correction to the raw images.
The total number of training samples per actor in this multiview data is $\sim$3k, covering the frontal head and the sides. 
For the experiments that show full $360 ^{\circ}$ head avatar reconstructions (see \Cref{fig:360}), we use $\sim$12k samples captured from 26 cameras from the Multiface v2 dataset which also covers the back of the head.
For additional comparisons against the baselines that do not handle the back of the head, we sample $11$ frontal cameras from the dataset (see suppl. doc.).

\paragraph{Metrics}
To quantitatively evaluate our method, we perform self-reenactment on the test data. 
We use the pixel-wise L2 reconstruction error, the peak signal-to-noise ratio (PSNR), structure similarity (SSIM), and the learned perceptual image patch similarity (LPIPS) as image generation metrics.

\subsection{Comparison to State of the Art}
\label{sec:comparisonsota}

In \Cref{tab:monocularsotacomparison} and \Cref{fig:novelexpressionmonocular}, we show a quantitative and qualitative comparison to the state-of-the-art monocular head reconstruction methods.
As can be seen in \Cref{tab:monocularsotacomparison}, our method produces the best perceptual image quality metrics, as well as pixel-based reconstruction errors.
As our model is trained without the need of facial expression supervision, the generated image quality is sharp and able to reproduce details like teeth, eyes, and thin structures like glasses-frames and hairs (see \Cref{fig:novelexpressionmonocular}).
The baselines tend to produce blurry appearances, as the facial expression tracking yields inconsistent training data, especially for side views.
%

% Table
\begin{table}[t]%
    \centering
    \begin{minipage}{\columnwidth}
    \begin{tabular}{lllll}
      \toprule
      Method   & MSE $\downarrow$ & PSNR $\uparrow$ & SSIM $\uparrow$ & LPIPS $\downarrow$ \\ 
      \midrule
      IMavatar \cite{Zheng:CVPR:2022}  & 0.0031 & 25.88 & 0.92 & 0.10 \\
      Nerface \cite{Gafni_2021_CVPR}     & 0.0024 & 27.07 & \textbf{0.93} & 0.11 \\
      INSTA  \cite{zielonka2023instant}   & 0.0046 & 23.60 & 0.92 & 0.10 \\
      \midrule
      Our Method    &   \textbf{0.0023} & \textbf{27.44} & 0.91 & \textbf{0.06} \\
      \bottomrule
    \end{tabular}
    \end{minipage}
    \caption{
    Quantitative evaluation based on 4 sequences from NeRFace \cite{Gafni_2021_CVPR} and INSTA \cite{zielonka2023instant}. }
    \label{tab:monocularsotacomparison}
\end{table}%

%%%%%%%%%%%%%%%%%%%%%%%%%%%%%%%%%%%%%%%

\subsection{Novel View \& Expression Synthesis}
\label{sec:360}
In \Cref{fig:360}, we show novel viewpoint synthesis for full-head avatar models which are trained on the multi-view Multiface dataset~\cite{wuu2022multiface}.
Our model is able to reconstruct the entire head, including the back of the head.
In the suppl. doc., we show an additional comparison on this data, where we adapt INSTA~\cite{zielonka2023instant} to use multi-view data. However, it is not able to capture the same level of detail as our proposed method.

\begin{figure}[t]
    \centering
    \includegraphics[keepaspectratio=true, width=\linewidth, height=0.6\linewidth]{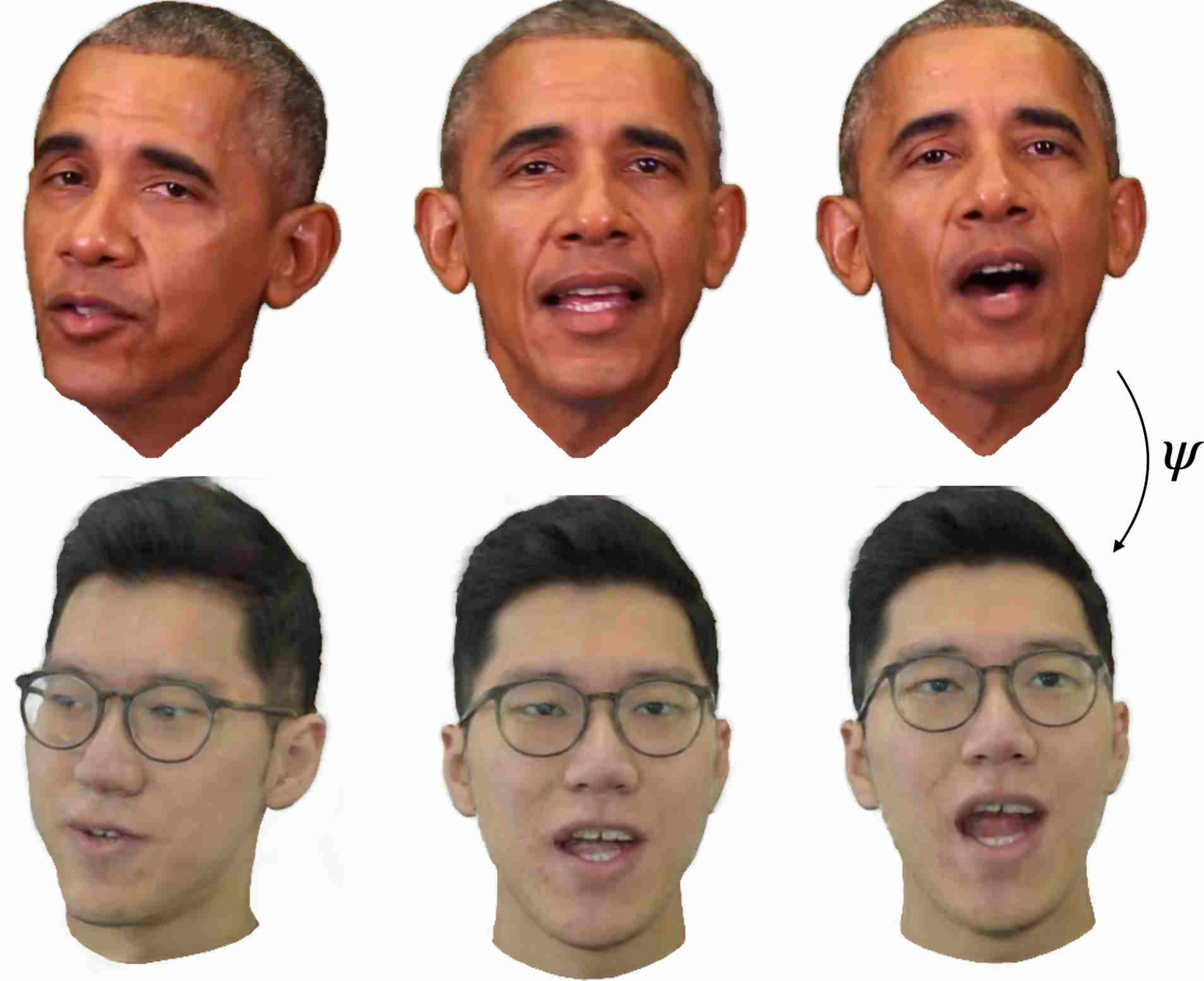}
    %\vspace{-0.4cm}
    \caption{
    Our 3D appearance models are controlled via 3DMM expression parameters, allowing for facial expression transfer, where the expressions of one person are applied to the avatar of another.
    }
    \label{fig:expressiontransfer}
    %\vspace{-0.4cm}
\end{figure}

Our method also allows us to transfer facial expression coefficients from one actor to another.
We demonstrate this facial expression transfer in  \Cref{fig:expressiontransfer}.

\subsection{Ablation Studies}
\label{sec:ablation}
\begin{figure}[t]
    \centering
    \includegraphics[width=0.8\linewidth]{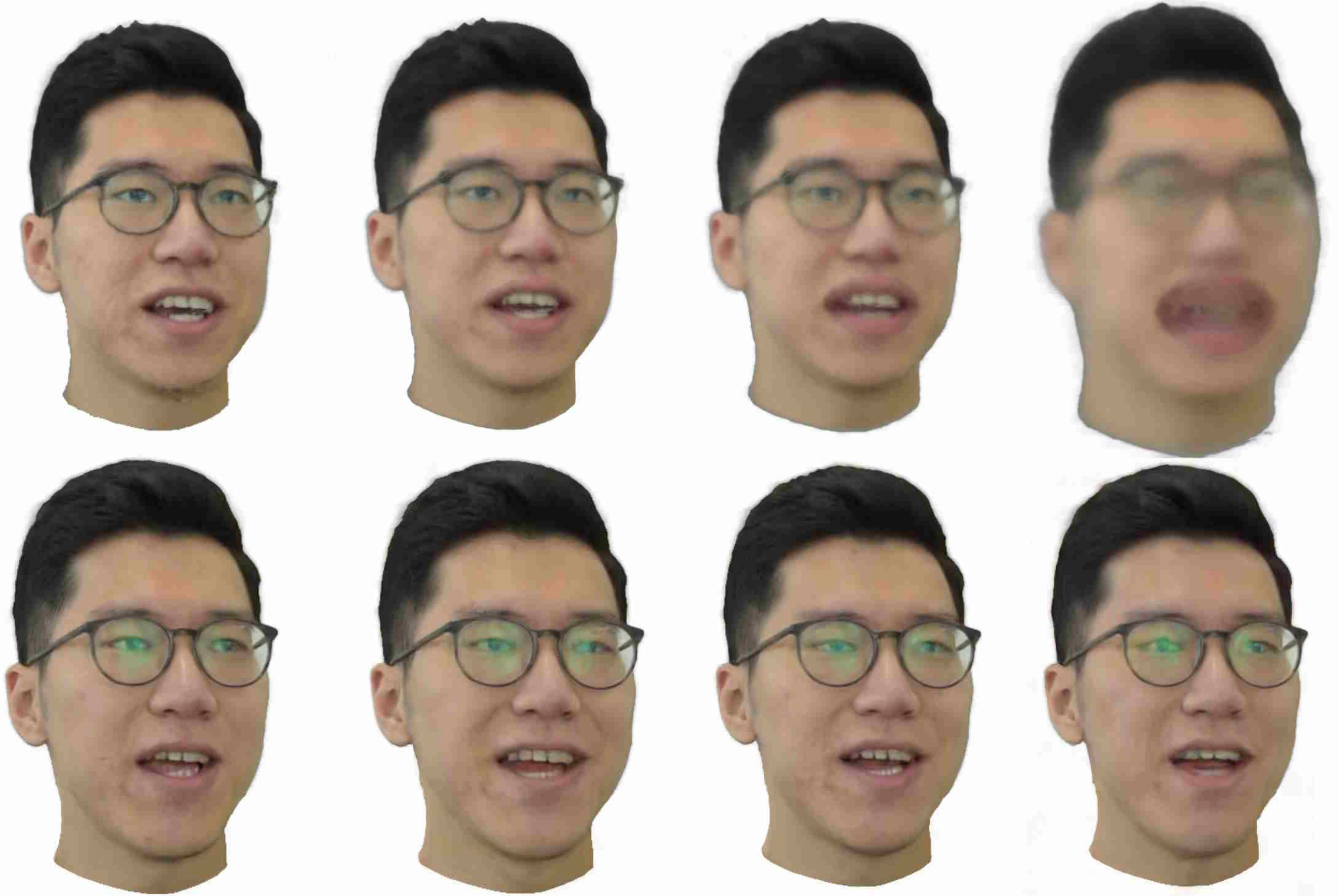}
    \text{\footnotesize No noise \hspace{1.0cm} 1mm  \hspace{1.0cm} 2mm  \hspace{1.0cm} 5mm  } 
    \caption{
    With an increasing noise level on translation, our method (second row) degrades gracefully and is still able to produce good appearances at a noise level of 5mm.
    In contrast, INSTA \cite{zielonka2023instant} (first row) heavily depends on precise face and camera pose tracking and averages the facial texture, leading to blurry results.
    }
    \label{fig:noise-exp}
\end{figure}

\paragraph{Robustness to imperfect camera poses}
To train our appearance model, we rely on paired input data of RGB images and camera poses.
We evaluate our model regarding noisy camera estimates and compare it to the state-of-the-art method, INSTA~\cite{zielonka2023instant}.
Specifically, we train appearance models where the camera poses are corrupted with increasing noise levels.
Both, INSTA and our method get the same camera poses as input~\cite{Zielonka2022TowardsMR}, while INSTA receives the facial expression as additional input (without noise).
We add translation noise to the cameras, using a Gaussian distribution with a mean $\mu$ of $0$ and varying $\sigma$ values (1mm, 2mm, and 5mm).
As can be seen in \Cref{fig:noise-exp}, despite the noise, our method is able to generate a good appearance model in comparison to INSTA, which gets increasingly blurry results.% (e.g., in the 2mm and 5mm case).

\begin{figure}[t]
    \centering
    \includegraphics[width=0.8\linewidth]{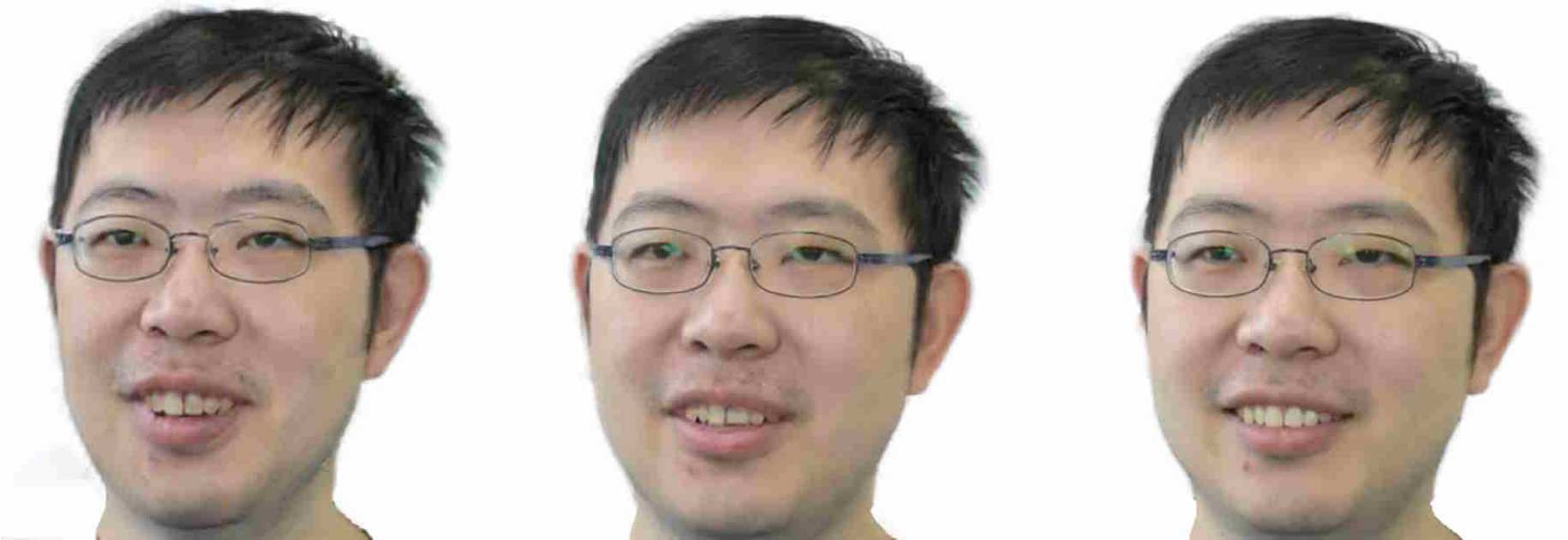}
    
    \text{\footnotesize \hspace{-0.8cm} w/ normalization \hspace{0.7cm} w/ Procrustes \hspace{1.0cm} GT}
    \caption{ Normalizing images based on landmarks enforces facial images to have the same distance between the eyes. However, this leads to distortions of the head when reconstructing a consistent 3D model, as the width of the head in the images is scaled differently for side and frontal views. 
    }
    \label{fig:ablation_norm}
   % \vspace{-0.5cm}
\end{figure}

\paragraph{Normalization of images}
EG3D is originally trained on FFHQ images, which are normalized based on facial landmarks.
These facial landmarks are only available for mostly frontal views, when the person is looking away from the camera the normalization cannot be applied, and the images have to be discarded.
Besides, normalizing images changes the geometry of the actor (i.e., narrowing face).
Instead of normalizing based on facial landmarks, we use Procrustes which allows us to preserve the identity of the actor (see \Cref{fig:ablation_norm}) and to use images from the back (see \Cref{fig:360}).

\section{Discussion}

Our proposed method is capable of producing highly realistic animatable $360^{\circ}$ head avatars without the need of facial expression tracking in the input data.
However, the method takes about $6-7$h to train on 8 NVIDIA A100-40GB GPUs.
As we are bound to the observed facial expression appearances spanned by the input data, our method cannot extrapolate to out-of-distribution expressions.
This is also a limitation of other state-of-the-art methods~\cite{Gafni_2021_CVPR,zielonka2023instant,grassal2022neural}, including methods like IMavatar~\cite{Zheng:CVPR:2022} which can deform the geometry to unseen expressions, but distorts the color appearance (e.g., stretching of teeth).

\section{Conclusion}
GAN-Avatar is a person-specific controllable head avatar generation method that does not require facial expression tracking (hard) of the training data.
Instead of learning a neural appearance layer on top of a mesh, we leverage a 3D-aware GAN to learn the facial appearance of the subject.
We can train this model on images of the entire head, including the back of the head, to get a high-quality $360^{\circ}$ head avatar.
To control this appearance model, we learn a mapping from classical facial blend shape parameters to the latent space of the 3D-aware GAN model.
As we have shown, our proposed method produces sharp and detailed imagery for novel expressions as well as novel viewpoints.
Our idea of tracker-free appearance learning with 3D-GANs, combined with the controllability of classical facial blendshape models does not suffer from facial expression tracking failures in the input data, and, thus, is a step towards high-quality digital doubles from commodity hardware.

\paragraph{Acknowledgement} 
We thank Balamurugan Thambiraja for his help with the video recording, Riccardo Marin and Ilya Petrov for proofreading, and all participants of
the study.
The authors thank the International Max Planck Research School for Intelligent Systems (IMPRS-IS) for supporting BK and WZ. JT is supported by Microsoft and Google research gift funds.
This work was supported by the German Federal Ministry of Education and Research (BMBF): Tübingen AI Center, FKZ: 01IS18039A. GPM is a member of the ML Cluster of Excellence, EXC 2064/1 – Project 390727645, and is supported by the Carl Zeiss Foundation.

{
    \small
    \bibliographystyle{ieeenat_fullname}
    \bibliography{main}
}

\clearpage
\appendix
\clearpage
\setcounter{page}{1}
\maketitlesupplementary
In this supplementary document, we provide additional ablation studies in \Cref{sec:additionalablation} and further comparison on the multi-view data using Multiface~\cite{wuu2022multiface} in \Cref{sec:mvcomparisonmultiface}. 
Moreover, we include additional experiments using Colmap in \Cref{sec:colmap}.

\section{Additional Ablation Studies}
\label{sec:additionalablation}

\begin{figure}[h]
    \centering
    \includegraphics[keepaspectratio=true, width=\linewidth, height=\linewidth]{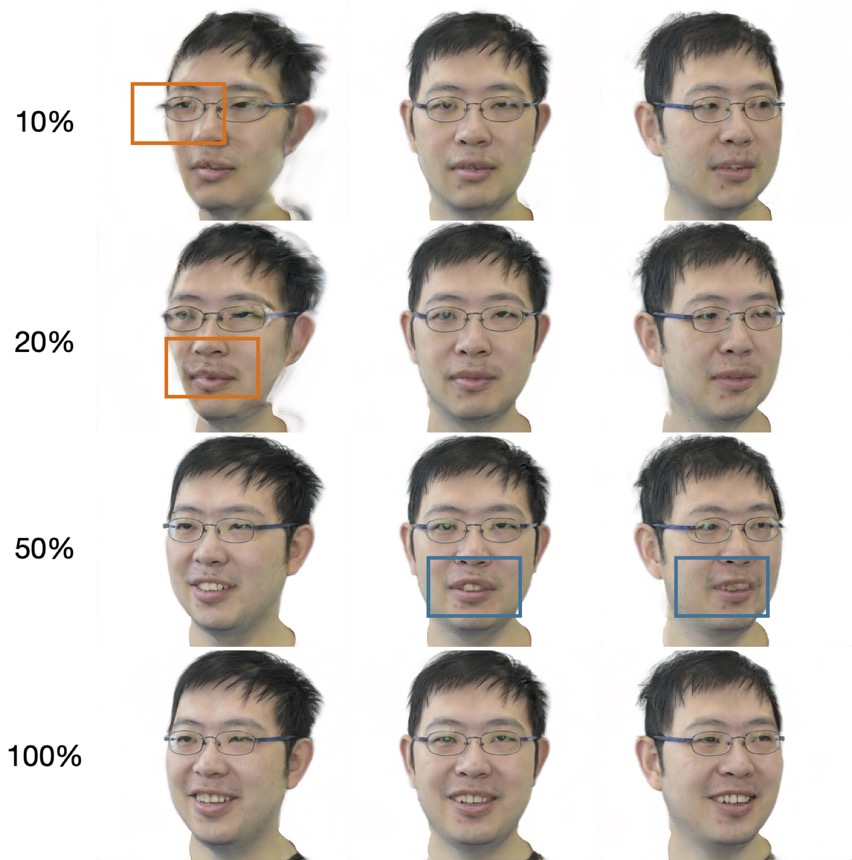}
    
    \caption{ Effect of the training data corpus size on the image quality.
    With a smaller dataset, expression inconsistencies between different camera poses occur.
    100\% corresponds to $\sim$3k RGB images.
    }
    \label{fig:synthesis_samples_ablation}
\end{figure}

\paragraph{Multi-view Consistency}
Our method is dependent on the training corpus size.
We assume to have the same training corpus size as the baseline methods, which typically require about 2-3min of monocular video data.
Using more samples with different camera views improves the consistency of the expressions from different angles and the image quality, as shown in \Cref{fig:synthesis_samples_ablation}.

\begin{figure}[h]
    \centering
    \includegraphics[width=1\linewidth]{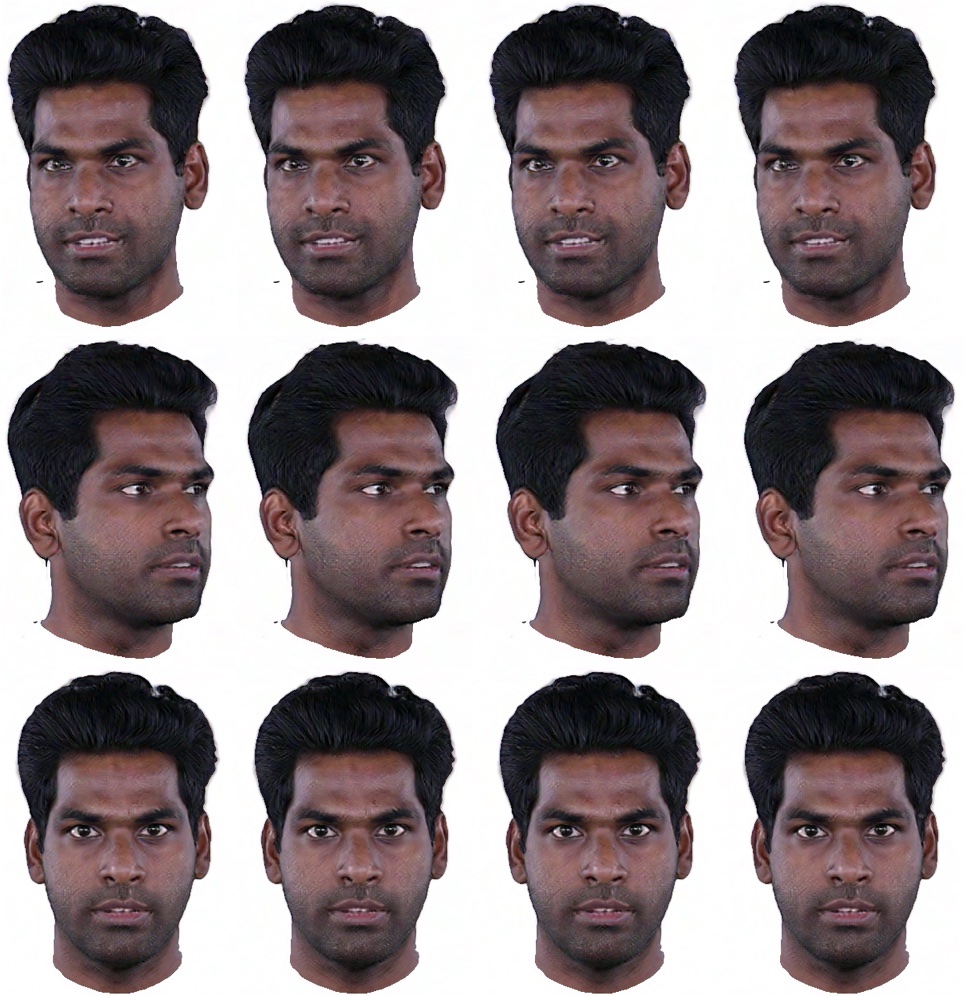}  
    \caption{
    The appearance model that does not utilize pre-training lacks expressiveness (i.e., the low number of different facial expressions).
    }
    \label{fig:no-pretrained-model}
\end{figure}

\begin{figure}[h]
    \centering
    \includegraphics[width=1\linewidth]{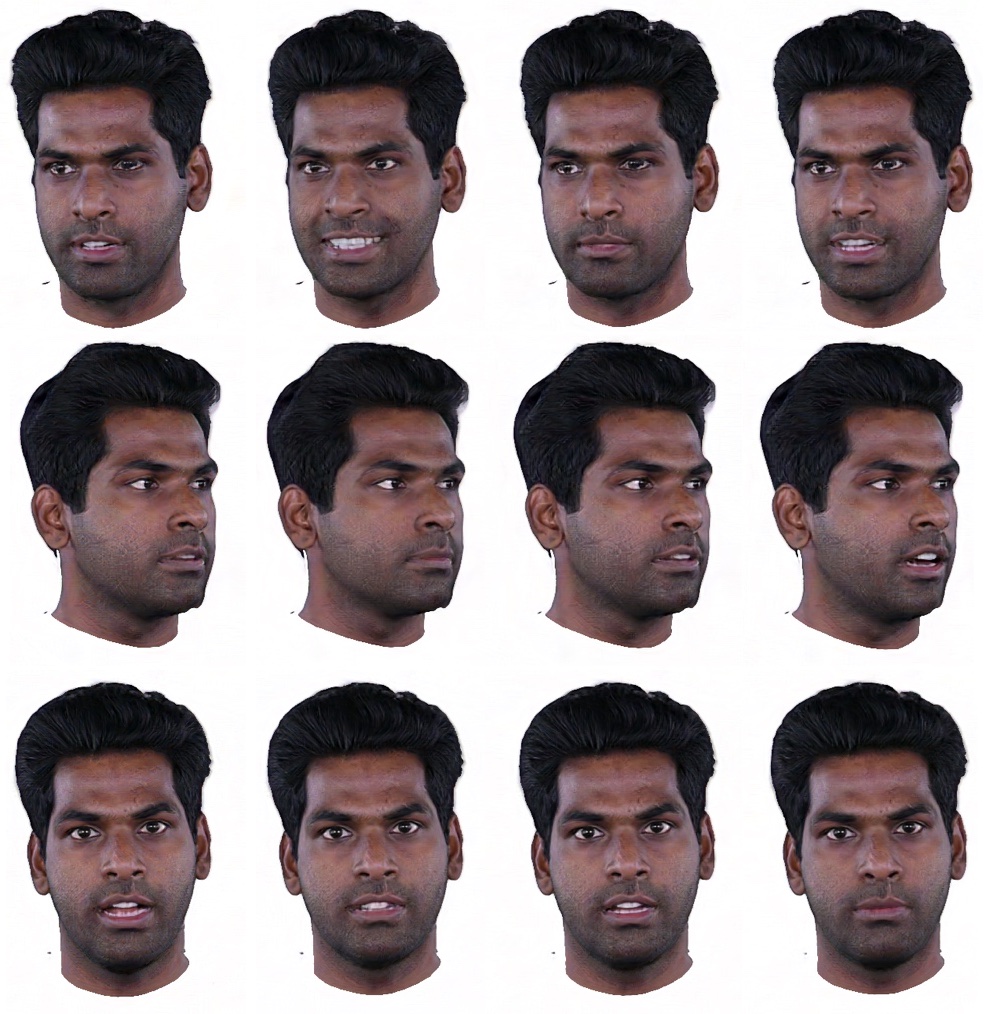}  
    \caption{
    Leveraging a pre-trained generative model trained on the FFHQ dataset helps us to converge faster and provides us with diverse expressions.
    }
    \label{fig:leverage-pretrained}
\end{figure}

\paragraph{Effect of Pre-training}
To train our appearance model, we leverage the pre-trained EG3D~\cite{chan2022efficient}.
To illustrate the effect of the pre-trained model, we train an additional appearance model without relying on any pre-training. 
We show the results in \Cref{fig:no-pretrained-model} and \Cref{fig:leverage-pretrained}.
Specifically, we train both models on 2 mins long videos and sample images from the respective models.
As can be seen, the network without pre-training generates similar-looking images in terms of expression and lacks diversity (see \Cref{fig:no-pretrained-model}), whereas the model that leverages pre-training produces a diverse set of facial expressions (see \Cref{fig:leverage-pretrained}).

\paragraph{Mapping Network -- Training Loss}
We use a photometric loss for training the expression mapping network.
An alternative is to directly train the network based on the predictions in latent space by measuring the distance between latent codes $\omega$ instead of using the photometric loss. 
As can be seen, the photometric loss performs slightly better than the loss in latent space, as shown in \Cref{tab:ablationsloss_w}.

\begin{table}[h]
    \centering
    \resizebox{\linewidth}{!}{
    \begin{tabular}{lllll}
      \toprule
      Method   & L2 $\downarrow$ & PSNR $\uparrow$ & SSIM $\uparrow$  & LPIPS $\downarrow$ \\ 
      \midrule
      Ours w/ $\omega$ loss      & 0.0025 & 26.11 & 0.68 & 0.14 \\
      Ours w/ img loss      & 0.0025 & 26.12 & 0.68 & 0.14 \\
      \bottomrule
    \end{tabular}
    }
    \caption{ Ablation study w.r.t. the training objective of the mapping network using the Multiface v2 dataset. \textit{$\omega$ loss} denotes the loss formulation in the latent space of StyleGAN2, while \textit{img loss} is the photometric loss used in our method.}
    \label{tab:ablationsloss_w}
\end{table}

%%%%%%%%%%%%%%%%%%%%%%%%%%%%%%%%%%%%%%%%%%%%%%%%%%%%%%

\section{Additional Comparison on Multiface Dataset}
\label{sec:mvcomparisonmultiface}

\begin{table}[t]%

    \centering
    \resizebox{\linewidth}{!}{
    \begin{tabular}{lllll}
      \toprule
      Method   & MSE $\downarrow$ & PSNR $\uparrow$ & SSIM $\uparrow$ & LPIPS $\downarrow$ \\ 
      \midrule
      INSTA-FL     & 0.0059 & 23.09 & 0.73 & 0.27 \\
      %INSTA-GT      & 0.0028 & 26.25 & 0.81 & 0.14 \\
      INSTA-MV     & 0.0027 & 26.51 & \textbf{0.82} & 0.13 \\
      \midrule
      Ours    &   \textbf{0.0026} & \textbf{26.60} & 0.76 & \textbf{0.10} \\
      \bottomrule
    \end{tabular}
    }
    \caption{
    Quantitative evaluation of novel expression synthesis using three unseen expression sequences from the Multiface dataset.
    In all metrics, our proposed method outperforms the multi-view baseline methods.
    }
    \label{tab:novelexpressionfrontal}
\end{table}%

\begin{figure*}[t]
    \centering\hspace{1cm}
    \includegraphics[width=0.78\linewidth]{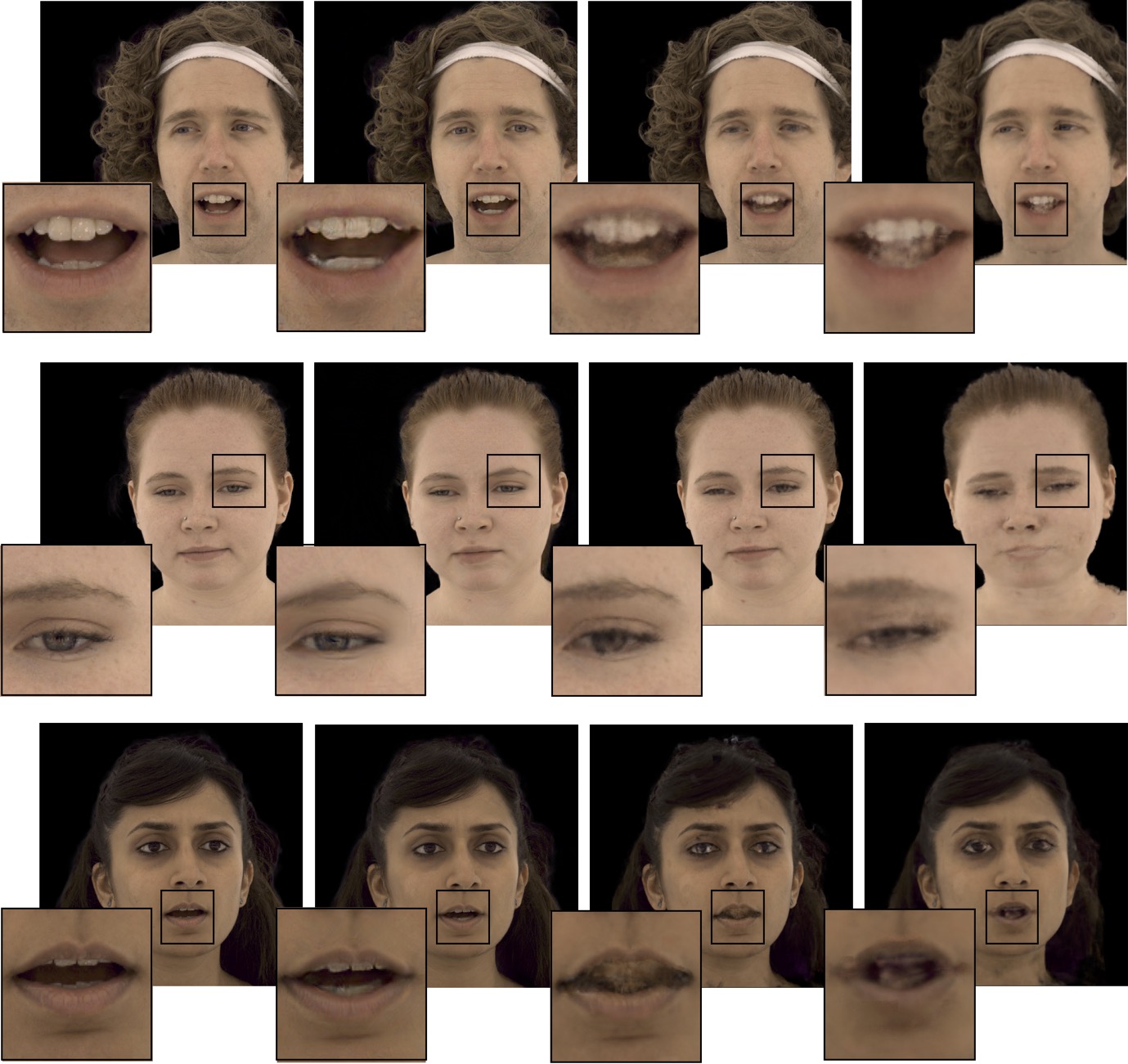} 
    \vspace{-0.75em}
    \newline
     \caption{
         Novel expression synthesis on the Multiface v2 dataset using the cameras from the frontal hemisphere.
         From left to right: ground truth (driving expression), our method, INSTA-MV and INSTA-FL.
         Notice the higher quality of our method in the teeth and eye regions.
         } 
    \label{fig:novelexpression}
\end{figure*}
As an additional baseline for the multi-view scenario~\cite{wuu2022multiface}, we modify the state-of-the-art method INSTA~\cite{zielonka2023instant}.
Specifically, we implemented two versions of INSTA, one which uses a multi-view FLAME tracking by adapting MICA~\cite{Zielonka2022TowardsMR} which we call INSTA-FL, and a second one which uses the production-ready motion capture of Laine et al.~\cite{laine2017tracking} which we call INSTA-MV.
For INSTA-MV, we use the production-ready motion capture provided by the Multiface dataset.
Note that this motion capturing is based on a person-specific template, including person-specific training of a tracking network.
Thus, it can not be easily applied to new subjects.
Both implementations allow us to use all multi-view images, including the back of the head.
Thus, INSTA-FL and INSTA-MV can also learn the back of the head.
We experimented with the loss formulation of INSTA and found that the usage of segmentation masks for $360^\circ$ avatar creation is leading to artifacts, as the face segmentation networks used in INSTA are not generalizing towards the back or the sides of the head.
Therefore, we disabled the segmentation-based loss together with the depth loss. We also double the number of iterations from $33k$ to $66k$.
We consider INSTA-MV as a strong baseline, as we provide production-ready tracking as input.
In contrast, our method only uses the images and corresponding camera distribution as input.
Note that other state-of-the-art methods like IMavatar~\cite{Zheng:CVPR:2022} behave similar to INSTA, however, are not trivially adaptable to the multi-view scenario, as segmentations and landmark networks fail to produce the required input.
We compare our method against INSTA-FL and INSTA-MV using sequences from the Multiface dataset with the v2 cameras, where the whole frontal head is covered (see \Cref{fig:novelexpression}).
Given an unseen test sequence of an actor, we extract the expression parameters using Deep3DFace~\cite{Deng2019Accurate3F} and use our mapping network to generate the corresponding latent codes $\omega$ from the given expression codes and render the resulting faces under a novel view.
Our method can reproduce the facial expressions of the ground truth input image and generates sharper output images than the baselines, which is also confirmed by the quantitative evaluation in \Cref{tab:novelexpressionfrontal}.
This is remarkable, as our method does not require any facial expression tracking of the input data.
Especially, in the mouth region which changes the most during different expressions, our method achieves clearer details (e.g., teeth).
Also, one can see the importance of accurate tracking for methods like INSTA.
INSTA-MV which uses production-ready, personalized face tracking achieves better visual quality than the FLAME-tracking-based INSTA-FL.

\section{Complete Head Avatar Reconstruction from Monocular Data }
\label{sec:colmap}

\begin{figure*}[t]
\centering
    \includegraphics[width=0.75\linewidth]{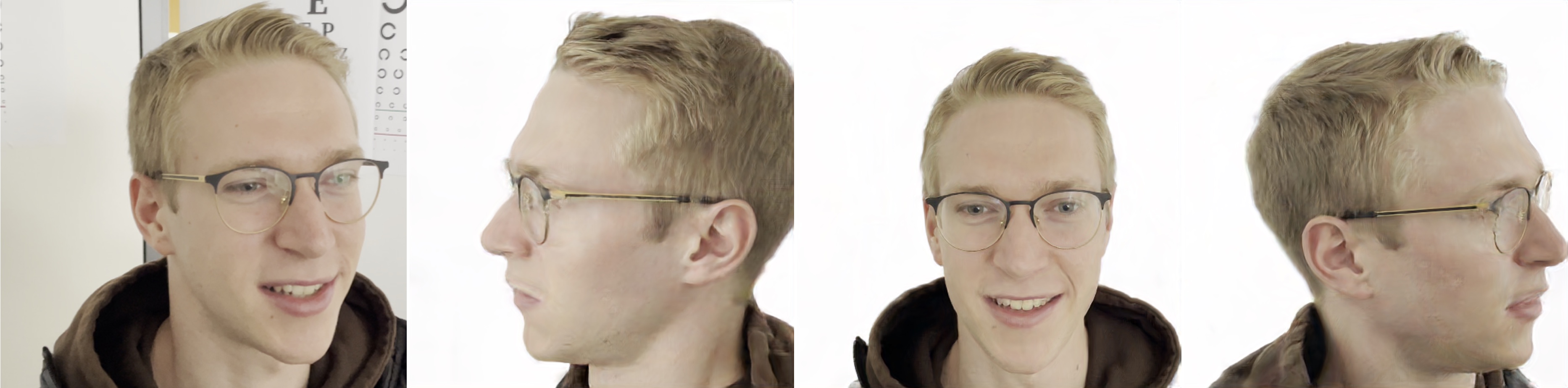} 
    \includegraphics[width=0.75\linewidth]{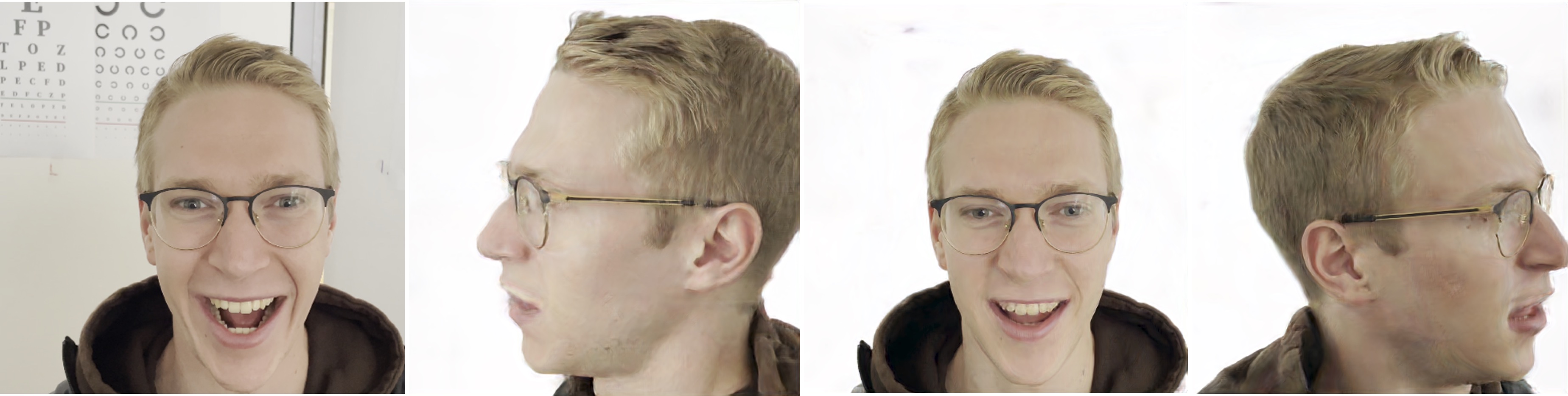} 
     \caption{
        Head avatar reconstruction from monocular data leveraging camera poses obtained via Colmap~\cite{schoenberger2016mvs, schoenberger2016sfm}. On the left, the ground truth is shown and next to it, novel-view point renderings of the 3D avatar. 
     } 
    \label{fig:drivingmica}
\end{figure*}

To further analyze the robustness of our method, we recorded a video that follows an oval trajectory and includes side views where landmark detectors fail.
This recording consists of 4537 frames, of which we use 4000 for training our appearance model.
To recover the camera poses, we use Colmap~\cite{schoenberger2016mvs, schoenberger2016sfm}. 
Specifically, we provide the RGB images and corresponding alpha masks obtained via video matting \cite{rvm} to Colmap's automatic sparse reconstruction method.
Using the resulting camera poses, we optimize our appearance model and learn a facial expression mapping network.
We use the last 500 frames of the recording as a test sequence, which is mostly frontal and is tracked with MICA \cite{Zielonka2022TowardsMR}. 
As shown in \Cref{fig:drivingmica}, one can see that our method can reconstruct a consistent 3D head avatar from this monocular data, including side views that cannot be tracked with a state-of-the-art facial expression tracking approach.

\end{document}